\documentclass[lettersize,journal]{IEEEtran}
\usepackage{amsmath,amsfonts}
\pdfoutput=1
\usepackage{algorithmic}
\usepackage{algorithm}
\usepackage{array}
\usepackage[caption=false,font=normalsize,labelfont=sf,textfont=sf]{subfig}
\usepackage{textcomp}
\usepackage{stfloats}
\usepackage{url}
\usepackage{verbatim}
\usepackage{graphicx}
\usepackage{hyperref}
\usepackage{cite}
\usepackage{multirow}
\usepackage{threeparttable}
\usepackage{booktabs}
\usepackage{color} 
\usepackage{amssymb}
\usepackage{amsmath}
\begin{document}

\title{Low-light Stereo Image Enhancement and De-noising in the Low-frequency Information Enhanced Image Space}

\author{\IEEEauthorblockN{Minghua Zhao\textsuperscript{1},Xiangdong Qin\textsuperscript{1},Shuangli Du\textsuperscript{1,\dag},Xuefei Bai\textsuperscript{1},Jiahao Lyu\textsuperscript{1},Yiguang Liu\textsuperscript{2}}
	\\
	\IEEEauthorblockA{\textit{1. Shaanxi Key Laboratory for Network Computing and Security Technology, School of Computer Science and Engineering, Xi'an University
			of Technology, Xi'an, China}
	\\
	\IEEEauthorblockA{\textit{2. College of Computer Science, Sichuan University, Chengdu, China}}
	\\
	\IEEEauthorblockA{\dag	Correspondence:\href{mailto:dusl@xaut.edu.cn}{dusl@xaut.edu.cn}}
}
}

\maketitle

\begin{abstract}
Unlike single image task, stereo image enhancement can use another view information, and its key stage is how to perform cross-view feature interaction to extract useful information from another view. However,  complex noise in low-light image  and its impact on subsequent feature encoding and interaction are ignored by the existing methods. In this paper, a method is proposed to perform enhancement and de-noising simultaneously. First, to reduce unwanted noise interference, a low-frequency information enhanced module (IEM) is proposed to suppress noise and produce a new image space. Additionally, a cross-channel and spatial context information mining module (CSM) is proposed to encode long-range spatial dependencies and to enhance inter-channel feature interaction. Relying on CSM, an encoder-decoder structure is constructed, incorporating cross-view and cross-scale feature interactions to perform enhancement in the new image space. Finally, the network is trained with the constraints of both spatial and frequency domain losses. Extensive experiments on both synthesized and real datasets show that our method obtains better detail recovery and noise removal compared with state-of-the-art methods. In addition, a real stereo image enhancement dataset is captured with stereo camera ZED2. The code and dataset are publicly available at: https://www.github.com/noportraits/LFENet.
\end{abstract}


\section{Introduction}
\IEEEPARstart{T}{he} application scenarios of binocular vision have become increasingly widespread, from the earliest robot vision to the current fields of autonomous driving, medical imaging, etc.. However, stereo images captured in a low-light environment often suffer from complex degradation, including low brightness, low contrast, color distortion, and noise, resulting in unreliable scene depth prediction.

At the beginning, researchers use single view enhancement technology to restore low-light stereo images, such as histogram equalization technology \cite{HE1}, \cite{HE2}, \cite{HE3}, Retinex-based methods \cite{retinex1}, \cite{retinex2}, \cite{retinex3}, \cite{retinex4}, \cite{retinex5} and deep learning based approaches \cite{dl4}, \cite{dl5}, \cite{dl6}, \cite{mirnet}, \cite{llformer}, \cite{LA}, \cite{zero}, \cite{retinexnet}, \cite{legan}, \cite{dl7}. Retinex-based technologies often decompose the input image into reﬂectance map and illumination map to describe image detail and  brightness information respectively. This allows to decouple enhancement task into two sub-tasks: adjusting brightness with illumination map, and suppressing noise with reﬂectance map \cite{retinex1}, \cite{retinex2}, \cite{retinex3}, \cite{retinex4}, \cite{retinex5}.

More recently, research in this area has begun to focus on stereo image enhancement. For stereo image pair, the two views exhibit strong correlation. Recent research has started to consider and utilize such correlation for stereo image enhancement \cite{stereo-resolution1}, \cite{stereo-resolution2}, \cite{stereo-resolution3}, \cite{stereo-resolution4}, \cite{PAM}, \cite{stereo-low1}, \cite{stereo-low2}, \cite{stereo-low3}, \cite{stereo-low4}, \cite{stereo-low5}. A key issue is how to perform cross-view interaction, incorporating the feature of the reference image to the target view. Parallax attention mechanism (PAM) is frequently used in cross-view interaction, which calculates the feature mapping matrix from the reference view to the target view. However, noise in low-light image will disturb the cross-view feature interaction, and tend to be amplified in the enhanced results, which is neglected by existing methods \cite{stereo-low1}, \cite{stereo-low2}, \cite{stereo-low3}, \cite{stereo-low4}, \cite{stereo-low5}.  

In this paper, we propose a new stereo image enhancement method, which can deal with illumination adjustment and denoising simultaneously. First, to reduce noise impact on the following feature encoding and interaction, instead of using an independent denoising module, we try to search a new image space to perform enhancement task. This means our method includes two stages. The first one is the image space searching stage, where low-frequency information is integrated into the original low-light image to reduce noise impact on subsequent feature encoding. The second stage is image enhancement. This framework allows to suppress noise in a two-stage manner. Second,  in addition to mining spatial information, dynamic feature channel interaction can suppress the unimportant noise channels and enhance important channels. For this, a cross-channel and spatial context information mining module is proposed to encode long-range spatial dependencies and to enhance inter-channel feature interaction. The proposed feature extraction module can eﬀectively solve the problem of information loss caused by the increase of network layers. In summary, the contribution of this paper are as follows:

(1)  We propose a new stereo image enhancement method, which can perform brightness improvement and denoising simultaneously. On the Holopix50, Flickr2014, and Kitti2015 datasets, it achieves SOTA performance in comparison with existing methods.

(2) To handle noise and reduce its impact on feature encoding and interaction,  a low-frequency information enhanced module (IEM) is proposed to search a new image space, where noise is suppressed. To the best of our knowledge, this is the first work attempt to perform enhancement task in a new image space.

(3) A feature extraction module, named cross-channel and spatial context information mining module (CSM) is proposed. CSM utilizes large convolution kernels, channel attention mechanisms and simple gate structure to encode long-range spatial dependencies and to enhance inter-channel feature interaction. It shows strong feature representation ability.

(4) We capture a real low-light stereo image enhancement dataset with a Zed2 camera, including 177 normal/low-light stereo image pairs. The dataset can be used in training stage together with synthetic dataset to improve model robustness.

The remaining sections of the paper are organized as follows. Section II reviews some related works about low-light single view enhancement and stereo image restoration and enhancement problems. Section III describes the overall architecture of the proposed method and presents the details. The performance evaluation of the proposed method in comparison with other methods is shown in Section IV. Section V concludes the paper.
\section{Related Work}
In terms of data processed, low-light image enhancement can be divided into three classes: single image enhancement, stereo image enhancement and video enhancement.
\subsection{Low-light Single View Enhancement}

\noindent \textbf{Traditional Methods: }The simplest low-light image enhancement method is histogram equalization \cite{HE1}, \cite{HE2}, \cite{HE3}, which enhance image contrast via adjusting image grayscale distribution. However, under extreme conditions, histogram equalization can easily introduce noise or cause overexposure and underexposure. Another popular and effective way to perform image enhancement is buillding retinex-based model \cite{retinex1}, \cite{retinex2}, \cite{retinex3}, \cite{retinex4}, \cite{retinex5}, usually decomposing the input image into two parts: the illumination part and the reﬂectance part. The illumination part is used to adjust the brightness, and the reﬂectance part is used to suppress noise and recover image details. At the beginning, researchs focus on developping Retinex decomposition model. Fu et al. \cite{retinex1} proposed a weighted variational decomposition model. Xu et al. \cite{retinex2} proposed a structure and texture aware Retinex (STAR) decomposition model with exponentiated local derivative constraints. Subsequently, researchs pay attention on noise suppression. Du et al. \cite{retinex3}, \cite{retinex4} enforced low-rank prior constraint on reflectance to remove noise. Differently, Hao et al. \cite{retinex5} utilized Gaussian total variation regularization and a TV-denoising term to remove noise. These noise suppression approaches are complex, time-consuming and tend to over-smooth image details.

\noindent \textbf{Learning-Based Methods: }In recent years, great performance improvement has been get by deep learning based appoaches. They can overcome the limitations of traditional methods by learning feature relationships in a large amount of images. Zamir et al. \cite{mirnet} pay attention on feature representation for real image restoration, and proposed a novel network architecture to maintain spatially precise high-resolution representations and to receive strong contextual information from the low-resolution representations. For ultra-high definition images, such 4K and 8K images, Wang et al. \cite{llformer} proposed a Transformer-based low-light enhancement method. For the problem of vehicle detection in low light conditions, Du et al. \cite{dl5} proposed a method that preserves the necessary details for detecting vehicles while enhancing the image. In order to solve the balance problem between overall brightness and local contrast enhancement, He el al. \cite{dl6} designed a multi-scale lighting adjustment and wavelet based noise cancellation network to effectively adjust lighting and remove noise.

Generally, low-light image enhancement task involves contrast enhancement, denoising, removing artifacts and color correction. Thus, it is reasonable and effective to decouple the task into multi-sub-tasks. Wu et al. \cite{retinexnet} and Liu et al. \cite{dl4} decoupled enhancement into two sub-tasks with Retinex model: adjusting brightness with illumination map, and suppressing noise with reﬂectance map. Similarly, Yang et al. \cite{LA} performed light adaptation in the low-frequency layer, and noise suppression or detail enhancement in high-frequency layer. 

The aforementioned approaches are the supervised solutions, requiring normal/low-light image pairs. It is very difficult or even impractical to capture normal/low-light image pairs simultaneously. So, some unsupervised and pseudo supervised solutions are introduced. Guo et al. \cite{zero} proposed a zero-reference method, which formulates light enhancement as a task of image-specific curve estimation with a deep network. Fu  et al. \cite{legan} proposed an unsupervised GAN-based approach incorporating attention module and identity invariant loss, which does not require paired data. For pseudo supervised low light image enhancement, Yu et al. \cite{dl7} used separable quadratic curves to generate pre enhanced images, and further explored the potential distribution of pseudo paired data through mutual learning strategies. Then, during the mutual learning process of two parallel networks, the enhanced images were obtained.
\subsection{Stereo Image Restoration and Enhancement}

Different from single image restoration approaches, stereo image restoration methods need to consider how to perform cross-view feature interaction. As a key branch, stereo image super-resolution has been widely investigated in recent years. Jeon et al. \cite{stereo-resolution1} directly utilized disparity priors to shift the dual view by several pixels to compensate for disparity and obtain complementary information. Wang et al. \cite{PAM} proposed an unsupervised parallax attention mechanism (PAM) to learn stereo correspondence, which has driven the development of stereo image super-resolution. Baed on PAM, Chu et al. \cite{stereo-resolution2} used cross attention to fuse features of left and right views without need of disparity prior measure stage. This model achieved competitive stereo image super-resolution by utilizing both intra-view and inter-view information. Lin et al. \cite{stereo-resolution3} combined PAM with Transformer to enhance features using self-attention mechanism. Zhu et al. \cite{stereo-resolution4} proposed a cascaded spatial perception module, combined with PAM, to reassign each position in the feature map based on its weight, making the captured cross-view information more eﬀective.

Recently, researchers have begun to pay attention on stereo low-light image enhancement. Hamed et al. \cite{stereo-low4} first applied CNN to the field. Huang et al. \cite{stereo-low2} first applied PAM to perfrom cross-view feature interaction in the field of stereo low-light image enhancement. They only perform PAM on the low-resolution features, and then upsample it to describe the PAM on the high-resolution features. Different from explicitly measuring feature correspondance between left and right views, Zhang et al. \cite{stereo-low1} used large convolution kernels to capture cross-view feature interaction at multiple scales. The above methods are challenging in dealing with stereo images with large disparity range. To handle this issue, Zheng et al. \cite{stereo-low3} perform PAM on multi-scale features to improve the reliability of cross-view feature interaction. To get real-time inference, Lamba et al. \cite{stereo-low5} proposed a lightweight network with Unet architecture. They process the left and right features at large-scale individually, because the two features do not align well due to large disparity. While, the left and right features at small-scale are extracted with a single module,  because the disparity between left and right features decreases and the network’s receptive field increases.

Different from stereo image super-resolution, stereo image pairs in low-light enhancement task suffer from complex degradarion, it would make the cross-view feature interaction highly uncertain, which is negneted by existing methods. In this paper, we try to search a new image space, and then in the new image space performing enhancement. 

\begin{figure*}[!t]\centering
	\includegraphics[width=6in]{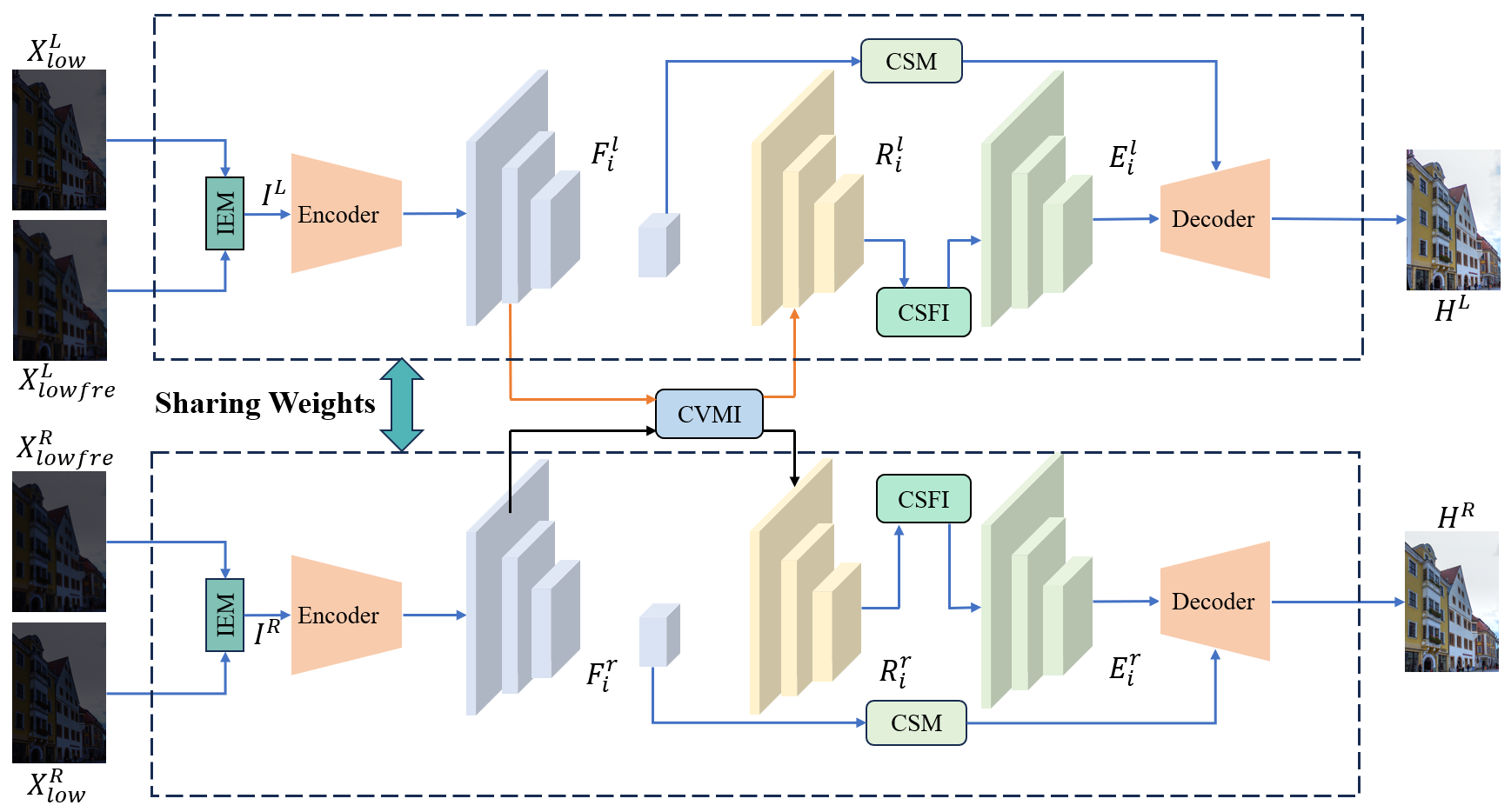}
	\caption{The overall framework of our proposed stereo image enhancement method, which contains two weights-shared branches to process left and right views respectively. The method includes three main modules, i.e., IEM, CVMI and CSFI. IEM takes in low-light images and its low-frequency part to suppress noise. CVMI performs cross-view feature interaction and CSFI performs interactions of multi-scale features of single view.}
	\label{pho:Overall Framework}
\end{figure*}
\section{Method}
In this section, we first introduce the overall architecture of the proposed method, followed by the implementation details of each module. At the end of this section, the used loss functions are discussed.

\subsection{Overall Architecture}

The proposed method overall architecture is shown in Figure \ref{pho:Overall Framework}. It includes three key modules, namely low-frequency information enhanced module, cross-view matching and interaction module, and cross-scale feature interaction module.

Unlike single image enhancement task, stereo image enhancement can use another view information via cross-view interaction. However, noise in low-light images would disturb the interaction. And such impact is always neglected by previous works \cite{stereo-low1}, \cite{stereo-low4}, \cite{stereo-low3}, \cite{stereo-low2}, \cite{stereo-low5}. In this paper, to reduce the noise impact on subsequent feature encoding, we propose a low-frequency information enhanced module(IEM). It incorporates low-frequency information into the original low-light images to produce a new image space, where noise is suppressed and low-frequency information is enhanced. Then image enhancement and de-nosing is performed in the new low-frequency enhanced image space with an encoder-decoder structure. The process can be described as:  
\begin{equation}
	\label{eq1}
	\left\{
	\begin{aligned}
		& F_{i}^{l},F_{i}^{r}=Encoder\left (  I^{L} \right ),Encoder\left (  I^{R} \right ) 
		\\
		&E_{i}^{l},E_{i}^{r}= \psi\left (  F_{i}^{l},F_{i}^{r} \right) 
		\\
		&H^{L},H^{R} =Decoder\left (  E_{i}^{l},E_{i}^{r} \right )  
	\end{aligned},
	\right.
\end{equation}
where \( I^{L}\) and \( I^{R}\) represent the left view and right view produced by the low-frequency information enhanced module; \(F_{i}^{l}\) and \(F_{i}^{r}\) are the encoded multi-scale features for single view image, i=1,2,3,4 denotes different scales; the function \(\psi\left (  \cdot \right ) \) includes two operations, i.e., cross-view and cross-scale feature interactions; \( E_i^l\) and \( E_i^r\) are the features for left view and right view obtained by \(\psi\left (  \cdot \right ) \) respectively; \(H^{L}\) and \(H^{R}\) represent the enhanced results for left and right views.

The cross-view matching and interaction module(CVMI) aims to find complementary cues from another view. The detail is introduced in Section \ref{section:cvmiandcsfi}(1). The cross-scale feature interaction module(CSFI) is used to improve each scale's feature expression ability via considering complementarity between multi-scales. It's detail is introduced in Section \ref{section:cvmiandcsfi}(2). Feature interactions are conducted on the three large scales. The 4-th scale feature does not join the interaction since its information is severely degraded.

\subsection{Low-Frequency Information Enhanced Module(IEM)}
\begin{figure}[!t]\centering
	\includegraphics[width=3.2in]{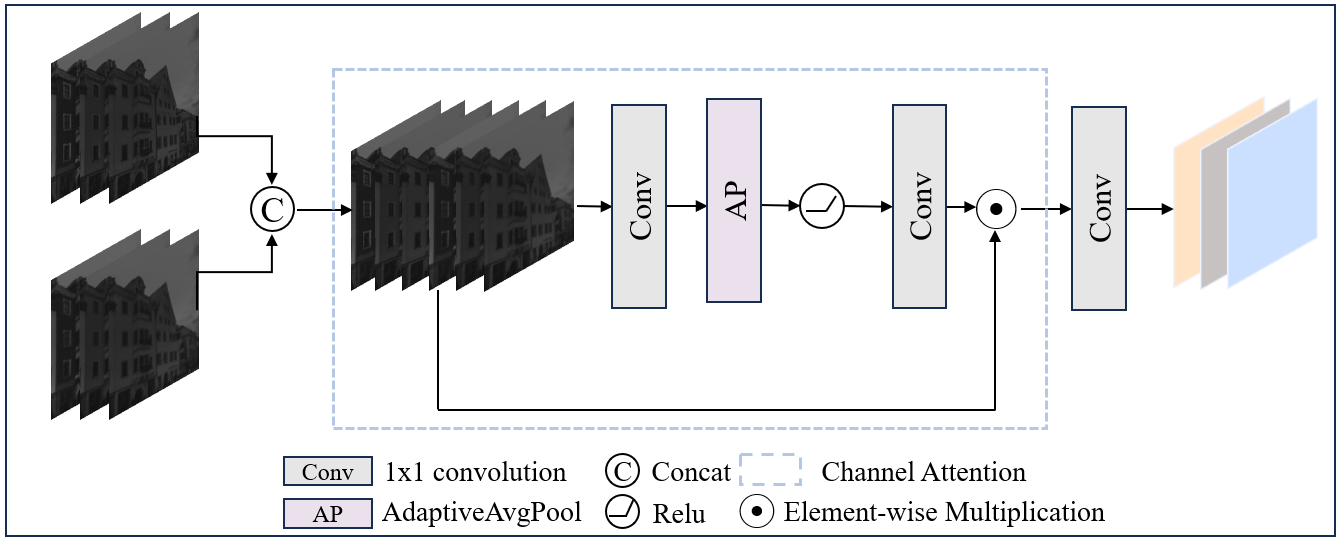}
	\caption{The detailed process of the low-frequency information enhanced module(IEM). The part highlighted by the dashed line is channel attention.}
	\label{pho:IEM}
\end{figure}

Generally, low-light images contain heavy and complex noise, which would disturb the cross-view information interaction, making it difficult to find valuable cues. In order to reduce the impact of noise on subsequent feature representation and mining, we perform enhancement in a low-frequency information enhanced image space. Specifically, we concatenate the input low-light image (\(X_{\text{low}}\)) and its low-frequency part (\(X_{\text{lowfre}}\)) along the channel dimension. It can be described as:
\begin{equation}
	\label{eq3}
	\left\{
	\begin{aligned}
		& I^{L}=CA\left (  Concat\left (  X_{\text{low}}^{L},X_{\text{lowfre}}^{L}\right ) \right )  
		\\
		& I^{R}=CA\left (  Concat\left (  X_{\text{low}}^{R},X_{\text{lowfre}}^{R}\right ) \right )  
	\end{aligned},
	\right.
\end{equation}
where \(Concat \left(\cdot \right) \) represents concatenation operation along channel dimension; \(CA\left (  \cdot \right ) \) stands for the Channel Attention (CA) mechanism\cite{CA}. The low-frequency part is extracted with side window filter\cite{lowfre}, which is very efficient. The detailed structure is illustrated in Figure \ref{pho:IEM}. Through the channel attention mechanism, the contribution of each channel can be adaptively adjusted. This enables the network to reduce noise impact with the help of low-frequency information, and to extract image details from the original low-light images, leading to positive contributions to subsequent feature encoding and interactions.  
\subsection{Feature Encoding and Decoding}
\subsubsection{Encoder-Decoder Module}
\begin{figure*}[!t]\centering
	\includegraphics[width=6in, height=4.2in]{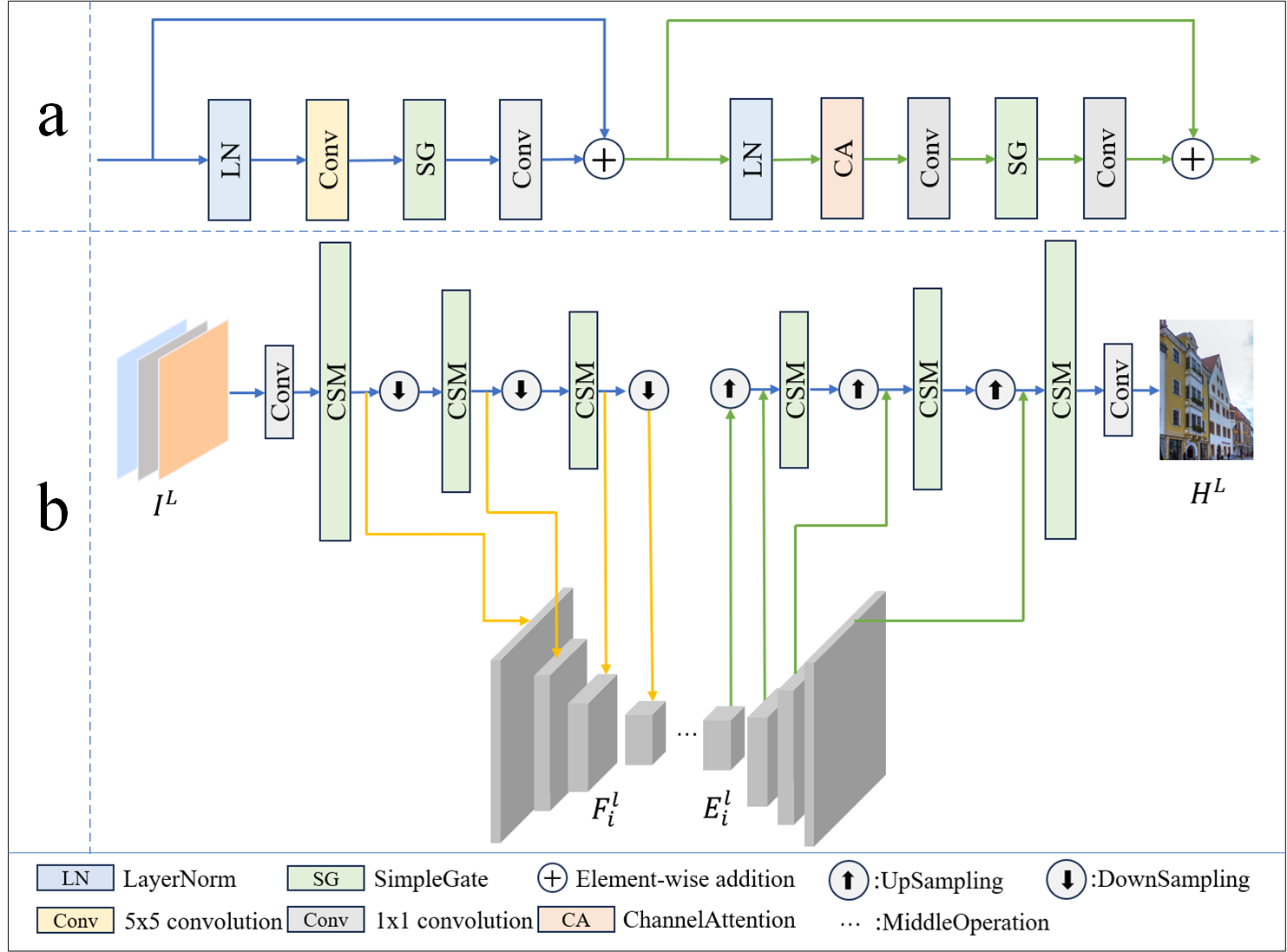}
	\caption{The detailed structure of feature encoding and decoding. (a): The structure of cross-channel and spatial context information mining module(CSM); (b): The main branch of encoder-decoder module.}
	\label{pho:EDandCSM}
\end{figure*}

Then, image enhancement and de-noising is performed in the low-frequency information enhanced image space with an encoder-decoder structure as shown in Figure \ref{pho:EDandCSM}(b). It includes channel expansion, three down-sampling operations, three up-sampling and channel recovery operations. Using 1 × 1 convolution expands the number of channels to 16, which helps to increase the network's expressive power. For each down-sampling, using a convolution operation with stride of 2 and kernel size of \(3 \times 3\) reduces feature size by half, while doubling feature channels, allowing the expansion of some spatial information into the channel dimension. The cross-view and cross-scale interactions are performed on the encoded multi-scale features here. For each up-sampling, using a deconvolution operation with a stride of 1 and kernel size of \(4 \times 4\) enlarges feature size by double, while reduces feature channels by half. Channel recovery is to reconstruct enhanced results from feature space. The encoder-decoder structure utilizes the proposed channel and spatial information mining module, called CSM, as the basic feature extraction method. Note that the encoder-decoder are used for both views, and both branches share parameters.

\subsubsection{Cross-Channel and Spatial Context Information Mining Module(CSM)}
As shown in Figure \ref{pho:EDandCSM}(a), CSM includes two stages, and each stage adopts residual structure. CSM focuses on mining long-range spatial dependencies and inter-channel information. Long-range spatial dependencies have been demonstrated to be highly eﬀective in image restoration tasks \cite{long1}, \cite{long2}, \cite{long3}. The frequently used methods for extracting long-range dependencies include transformer, large convolution, dilated convolution. Transformer has high computational resource requirements \cite{transformer1}, \cite{transformer2}, \cite{transformer3} and dilated convolution has the potential for information loss \cite{dc1}, \cite{dc2}. Some research \cite{large1}, \cite{large2}, \cite{large3} results suggest that large convolution can achieve well performance in low-level vision tasks. Thus, in stage one, we use large convolution kernels  (i.e., \(5 \times 5\))  to get long-range spatial context dependencies. To explore and emphasize the inter-channel information relationship, the channel attention (CA) mechanism\cite{CA} is used to dynamically decide the contribution of each channel. Then, to enhance inter-channel interaction, \(1 \times 1\) cross-channel convolution is used to increase and compress features channels, and SimpleGate mechanism \cite{simplegate} is adopted.

SimpleGate structure partitions feature  \(x \in\mathbb{R}^{C\times H\times W}\) into two parts (i.e.,  \( x_{1}, x_{2} \in \mathbb{R}^{\frac{C}{2}\times H\times W}\)) along the channel dimension and computes their dot product. Unlike commonly used ReLU and Sigmoid, our approach aims to minimize information loss during activation, as suggested by previous studies\cite{stereo-resolution2}, \cite{simplegate}. The specific implementation can be described as the following formula:
\begin{equation}
	\label{eq4}
	\left\{
	\begin{aligned}
		& x_{1},x_{2}=x\left [:\frac{C}{2},H,W  \right ]  ,x\left [\frac{C}{2}:,H,W  \right ]  
		\\
		& SG\left (x  \right ) =x_{1}\odot x_{2}
	\end{aligned},
	\right.
\end{equation}
where \(SG\left ( \cdot  \right ) \) represents SimpleGate; x represents the input features; \(\odot\) represents the element-wise multiplication.

\subsection{Cross-View and Cross-Scale Interaction Modules}
\label{section:cvmiandcsfi}
\subsubsection{Cross-View Matching and Interaction Module(CVMI)}

For stereo image pair, the two views exhibit strong correlation. Recent research has started to consider and utilizes such correlation for stereo image enhancement \cite{stereo-resolution1}, \cite{stereo-resolution2}, \cite{stereo-resolution3}, \cite{stereo-low3}, \cite{stereo-low4}. To exploit useful information from another view, we develop a cross-view matching and interaction module based on parallax-attention mechanism (PAM)\cite{PAM}. For stereo images, the corresponding pixel for a pixel in the left image only lies along its epipolar line in the right image. PAM considers epipolar constraints and uses geometry-aware matrix multiplication to calculate feature correlation between any two positions along the epipolar line, resulting in a parallax-attention map \(T^{r\rightarrow l}\) and \(T^{l\rightarrow r}\). The specific structure of CVMI is illustrated on the left side of Figure \ref{pho:CVMIandCSFI} and can be formulated as follows:
\begin{figure*}[!t]\centering
	\includegraphics[width=6.3in]{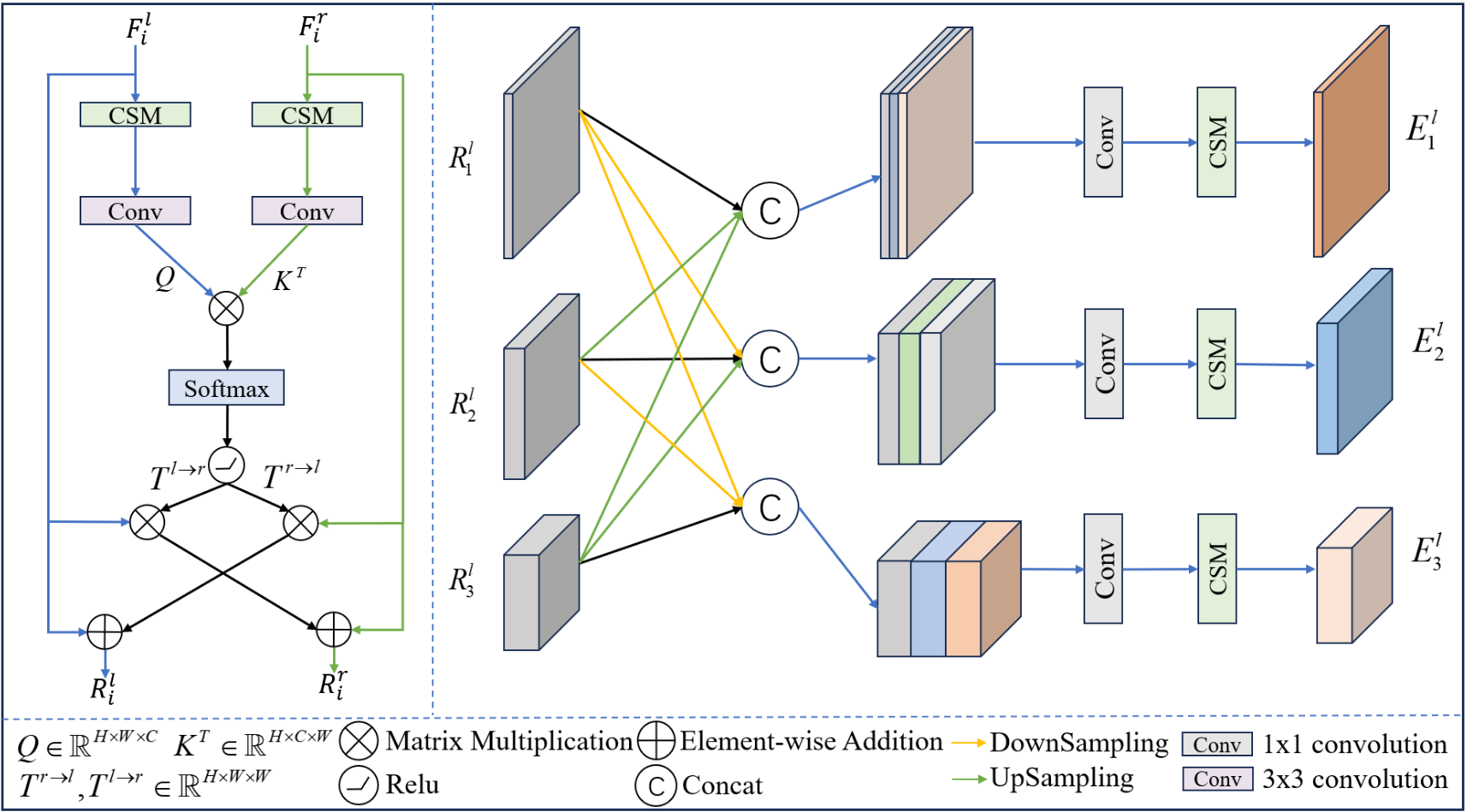}
	\caption{The left part is the details of Cross-View Matching and Interaction Module(CVMI), where only one scale of feature interaction is shown. The right part is the details of Cross-Scale Feature Interaction Module(CSFI).}
	\label{pho:CVMIandCSFI}
\end{figure*}

\begin{equation}
	\label{eq5}
	\left\{
	\begin{aligned}
		& R_{i}^{l},R_{i}^{r}=F_{i}^{l}+F_{i}^{r} \otimes T^{r\rightarrow l}, F_{i}^{r}+F_{i}^{l} \otimes T^{l\rightarrow r}
		\\
		& T^{r\rightarrow l} =Softmax(Q\otimes K^{T})
		\\
		& T^{l\rightarrow r} = Softmax(K\otimes Q^{T})
		\\
		& Q,K=Conv_{3}(CSM\left (F_{i}^{l} \right )),Conv_{3}(CSM\left (F_{i}^{r} \right ))
	\end{aligned},
	\right.
\end{equation}
where \(F_{i}^{l}\) and \(F_{i}^{r}\) are the encoded multi-scale features for single view image as given in Equation \ref{eq1}, \( i=1,2,3\). \(R_{i}^{l}\) and \(R_{i}^{r}\) are the multi-scale features for left and right view after cross-view interaction. \(Conv_{3}\) represents 3×3 convolutional. \(T^{l\rightarrow r}\) is the attention map from the left view feature to the right view feature. \(T^{r\rightarrow l}\) is the attention map from the right view feature to the left view feature. We perform feature correlation measure in a new feature space instead of using \(F_{i}^{l}\) and \(F_{i}^{r}\). Because there is a gap between the latent feature space learned for different vision tasks.
\subsubsection{Cross-Scale Feature Interaction Module(CSFI)}
Features at diﬀerent scales show diﬀerent representation capabilities and semantic information. To facilitate information exchange and integration among multi-scales, a cross-scale feature interaction module is utilized. More precisely, we upsample or downsample each scale, extending them to several other scales. The specific structure is illustrated on the right side of Figure \ref{pho:CVMIandCSFI} and can be formulated as follows:
\begin{equation}
	\label{eq7}
	\left\{
	\begin{aligned}
		& E_{1}^{l}=CSM\left ( Conv_{1}\left ( Concat\left (  R_{1}^{l},R_{2}^{l}\uparrow,R_{3}^{l}\uparrow\uparrow  \right )  \right )   \right )  
		\\
		& E_{2}^{l}=CSM\left ( Conv_{1}\left ( Concat\left (  R_{1}^{l}\downarrow ,R_{2}^{l},R_{3}^{l}\uparrow  \right )  \right )   \right )   
		\\
		& E_{3}^{l}=CSM\left ( Conv_{1}\left ( Concat\left (  R_{1}^{l}\downarrow\downarrow ,R_{2}^{l}\downarrow,R_{3}^{l}  \right )  \right )   \right )    
	\end{aligned},
	\right.
\end{equation}
where \(\uparrow\) represents upsampling, \(\downarrow\) represents downsampling, and \(Conv_{1}\) denotes a 1x1 convolution used for channel restoration, \(Concat \left(\cdot \right) \) represents a concatenation operation. Here we only show the processing of left view features, and we use the same operation for right view features.
\begin{table*}[h]
	\centering
	\caption{Evaluation results of each method on Holopix50, Flickr1024 and KITTI2015 datasets. The bold numbers represent the best performance. It can be observed that compared with other methods, our method shows the SOTA performance on all test datasets.}
	\begin{tablenotes}
		\item \small
	\end{tablenotes}
	\resizebox{7in}{0.8in}{
		\begin{tabular}{cccccccc}
			\midrule[1pt]
			&                         & \multicolumn{2}{c}{{Holopix50}} & \multicolumn{2}{c}{{Flickr2014}} & \multicolumn{2}{c}{{Kitti2015}} \\ \cline{3-8} 
			\multirow{-2}{*}{Methods} & \multirow{-2}{*}{Venue} & Left                     & Right                   & Left                    & Right                   & Left                      & Right                    \\
			\hline
			Zero-DCE \cite{zero}                  & CVPR 2020                         & 14.926/0.550             & 14.991/0.546            & 16.293/0.543            & 16.334/0.543            & 16.239/0.621              & 16.260/0.626             \\
			RetinexNet \cite{retinexnet}                & CVPR2022                          & 19.486/0.625             & 19.453/0.619            & 19.936/0.612            & 19.899/0.610            & 19.408/0.685              & 19.490/0.687             \\
			MIRNet \cite{mirnet}                    & ECCV 2020                         & 25.187/0.818             & 25.378/0.806            & 21.351/0.732            & 21.390/0.733            & 23.882/0.782              & 23.879/0.778              \\
			LA-Net \cite{LA}                    & IJCV 2023                         & 23.849/0.769             & 24.010/0.760            & 21.798/0.705            & 21.765/0.705            & 24.928/0.837              & 24.841/0.830             \\
			LLFormer \cite{llformer}                  & AAAI 2023                         & 25.255/0.812             & 25.113/0.808            & 25.812/0.803            & 25.834/0.804            & 32.017/0.906              & 31.867/0.903             \\
			DVE-Net \cite{stereo-low2}                    & TMM 2022                          & 25.245/0.822             & 25.202/0.813            & 26.112/0.810            & 26.156/0.810            & 31.863/0.906              & 31.803/0.904             \\
			DCI-Net \cite{stereo-low3}                    & MM 2023                         & 25.269/0.833             & 25.308/0.823            & 26.214/0.823            & 26.220/0.822            & 32.586/0.916             & 32.224/0.913             \\
			\hline
			Ours                      & -                         & \textbf{26.586/0.868}             & \textbf{26.342/0.860}            & \textbf{26.843/0.853}           & \textbf{26.859/0.853}            & \textbf{32.998/0.927}              & \textbf{32.937/0.926} \\ 
			\midrule[1pt]
		\end{tabular}
	}
	\label{tab:testdata}
\end{table*}
\subsection{Loss Functions}
The total loss function includes two parts, i.e., frequency domain loss \(\mathcal{L}_{fre} \) and spatial domain loss \(\mathcal{L}_{spa}\):
\begin{equation}
	\label{eq8}
	\mathcal{L}=\mathcal{L}_{fre}+\mathcal{L}_{spa} 
\end{equation}

By applying Fourier transform, we transform the enhanced image and ground truth image into the frequency domain, and then calculate the frequency domain \(\mathcal{L}_{fre}\) loss to measure their similarity. This component focuses on restoring image-level noise and details. It can be described by the following formula:
\begin{equation}
	\label{eq9}
	\begin{split}
		\mathcal{L}_{fre}&=\left \|  FFT\left ( H^{L} \right ) ,  FFT\left ( H_{GT}^{L} \right )  \right \| _{1} \\
		& + \left \|  FFT\left ( H^{R} \right ) ,  FFT\left ( H_{GT}^{R} \right )  \right \| _{1},
	\end{split}
\end{equation}
where \( H_{GT}^{L}\) and \( H_{GT}^{R}\) represent the ground truth of the left and right views in stereo image pairs; \(FFT\left ( \cdot\right )\) represents the fast fourier transform, \(\left \|  \cdot\right \| _{1}\) represents the \(\mathcal{L}_{1}\) loss function. 

On the other hand, considering spatial features such as contours and colors, we measure the pixel level diﬀerences between enhanced images and ground truth with structural similarity. It can be described by the following formula:
\begin{equation}
	\label{eq10}
	\begin{split}
		\mathcal{L}_{spa}&=1-SSIM\left ( H^{L},H_{GT}^{L} \right ) \\
		& +1-SSIM\left ( H^{R},H_{GT}^{R} \right ),
	\end{split}
\end{equation}
where \(SSIM\left ( \cdot\right )\) represents the structural similarity between two images. Subtracting it from 1 to obtain the restoration loss in space.
\begin{figure*}[h]\centering
	\includegraphics[width=5.4in]{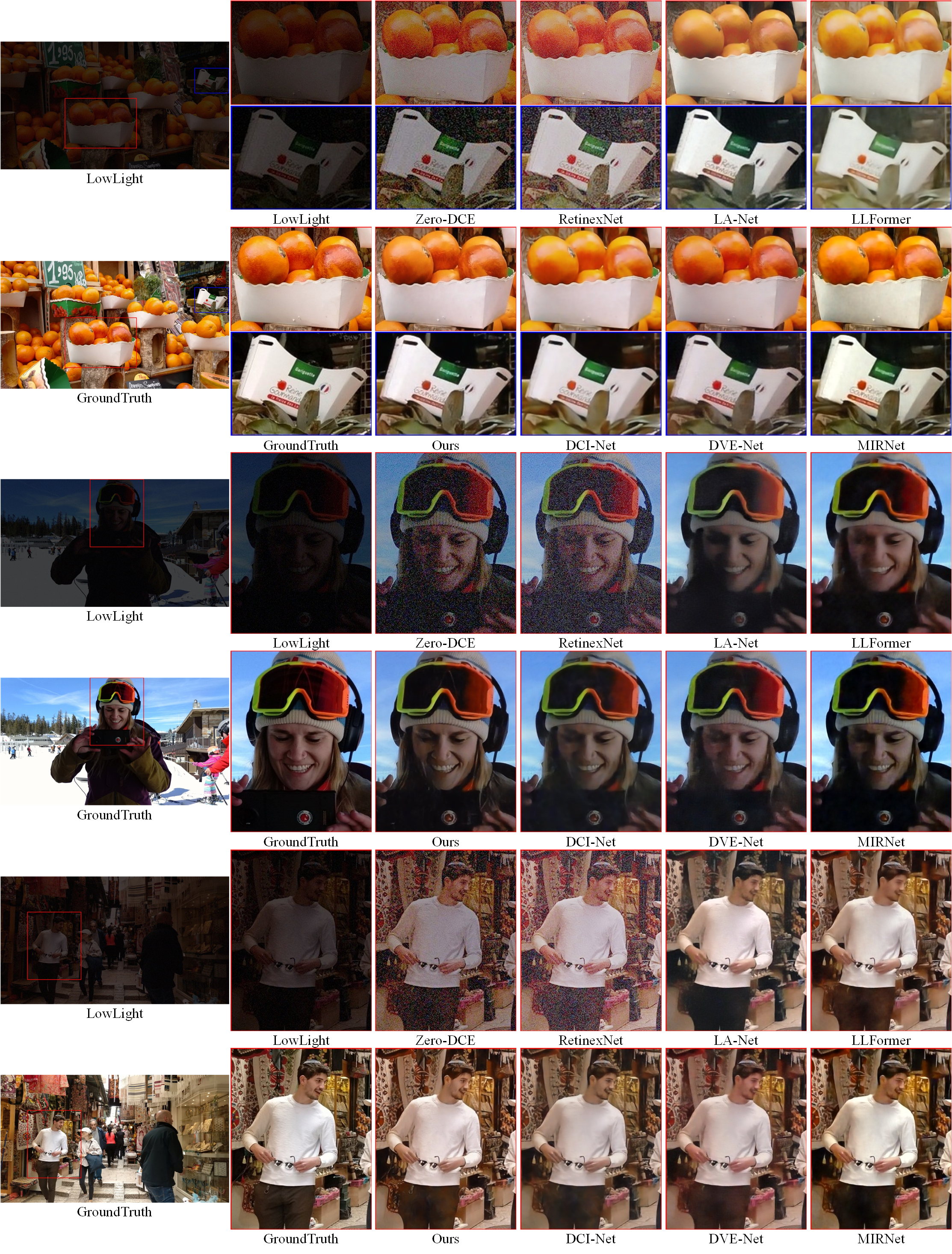}
	\caption{Visualization of the enhanced image of each method on Holopix50 dataset. It can be seen that compared with other methods, our method restores the texture and color information better.}
	\label{pho:holopix}
\end{figure*}
\begin{figure*}[!t]\centering
	\includegraphics[width=5.3in]{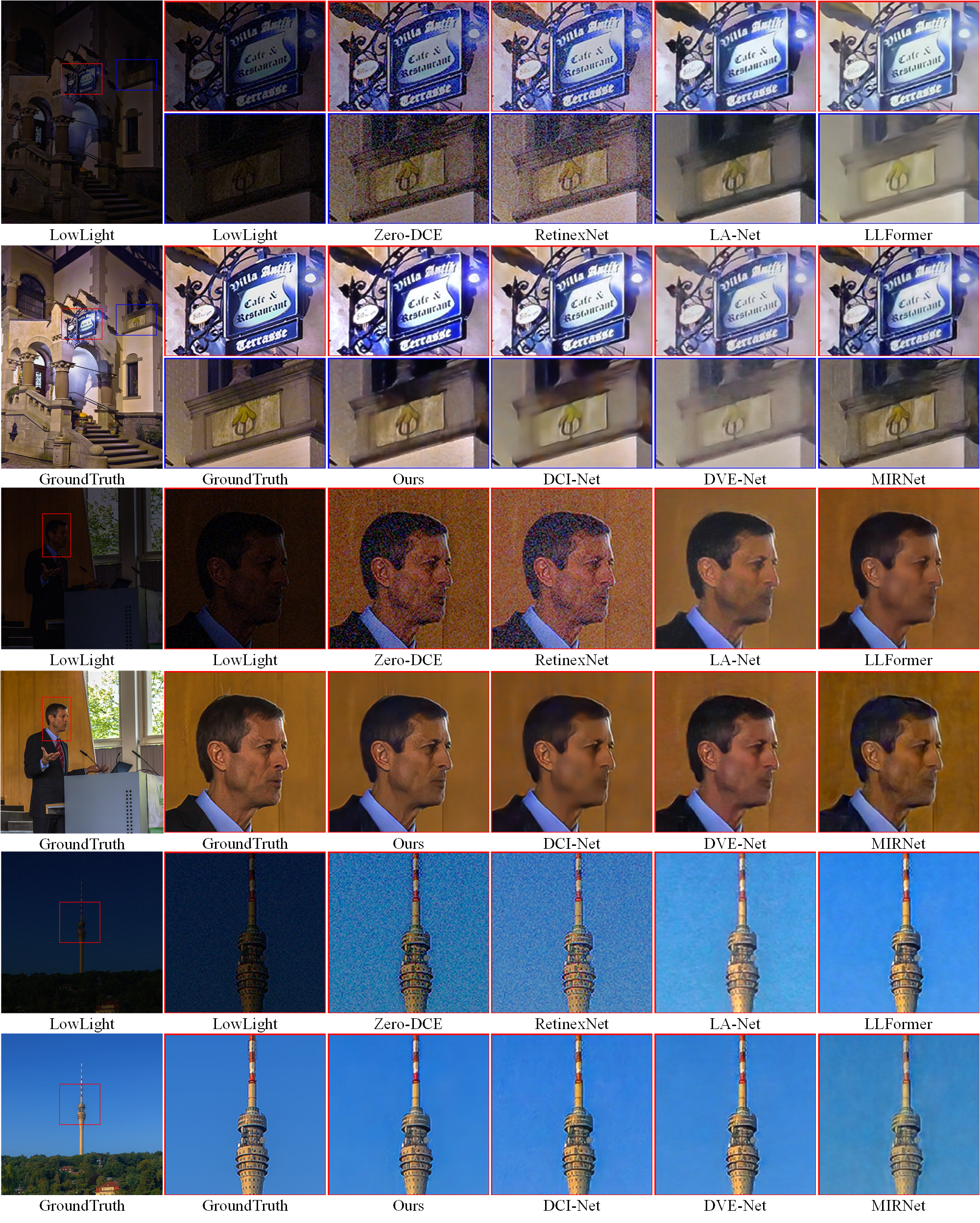}
	\caption{Visualization of the enhanced image of each method on Flickr2014 dataset. It can be seen that compared with other methods, our method effectively removes the influence of noise, at the same time, our method achieves the best details recovery.}
	\label{pho:flickr} 
\end{figure*}

\section{EXPERIMENTS}
In this section, we first introduce the dataset we use and the production process, then introduce our implementation details and comparison methods, and finally explain our experimental results and detailed analysis.
\subsection{Experiment Setting}

\subsubsection{Datasets}
The proposed stereo image enhancement method requires low/normal-light image pairs to train. However, the frequently used  stereo matching datasets do not have images obtained under low-light condition. So, we select some normal-light stereo image pairs to produce their synthetic low-light version. The selection is carried out based on subjective image quality assessment and two non-reference image quality indicators, i.e., NIQE and NIQMC \cite{NIQMC}. If \(NIQE < 2.5\) and \(NIQMC > 5\), the image is selected. To leverage the diversity of training data, we selected 1958 stereo images pairs to generate synthesized data, including 1189 images from Holopix50, 391 images from Flickr2014, and 378 images from Kitti2015. They are captured from diverse scenes. And 1197 stereo image pairs are used for training, 24 pairs are used for validation, and 737 pairs are used for test. Specifically, to generate synthetic low-light images, first we use gamma correction to lower the luminance of the images. It can be described by the following formula:
\begin{equation}
	\label{eq12}
	I_{out}=\beta \times \left ( \alpha \times I_{in}  \right )^{\gamma },   
\end{equation}
where \(I_{in}\) is the input image, and \(I_{out}\) is the obtained low-light image. To produce low-light images with different-level illumination, the value of parameters \(\alpha\), \(\beta\) and \(\gamma \) were selected randomly from two sets of parameters. The first set is defined as \(\alpha \sim U\left (  0.65,0.7\right ) ,\beta \sim U\left ( 0.65,0.7 \right ), \gamma \sim U\left ( 1.5,1.6 \right )\), and the second set is defined as \(\alpha \sim U\left (  0.8,0.85\right ) ,\beta \sim U\left ( 0.8,0.85 \right ), \gamma \sim U\left ( 3,3.2 \right )\), where \(U\) represents uniform distribution. The parameters value are selected with a ratio of 4:1 from the two sets.

Then we added noise using the method \cite{noise} to simulate noise distribution in real-world scenarios, where signal-dependent noise \(\sigma_{s} \sim U (0.09,0.1)\), and signal-independent noise \(\sigma_{c} \sim U(0.02,0.03)\).

In addition, we capture a real low-light stereo image enhancement dataset with a Zed2 camera, including 177 normal/low-light stereo image pairs. To the best our knowledge, it is the first real low-light stereo image enhancement dataset. The dataset can be used in training stage together with synthetic dataset to improve model robustness. We select 72 real stereo image pairs for training, and 105 pairs for test.
\begin{figure*}[t]\centering
	\includegraphics[width=6.5in, height=7in]{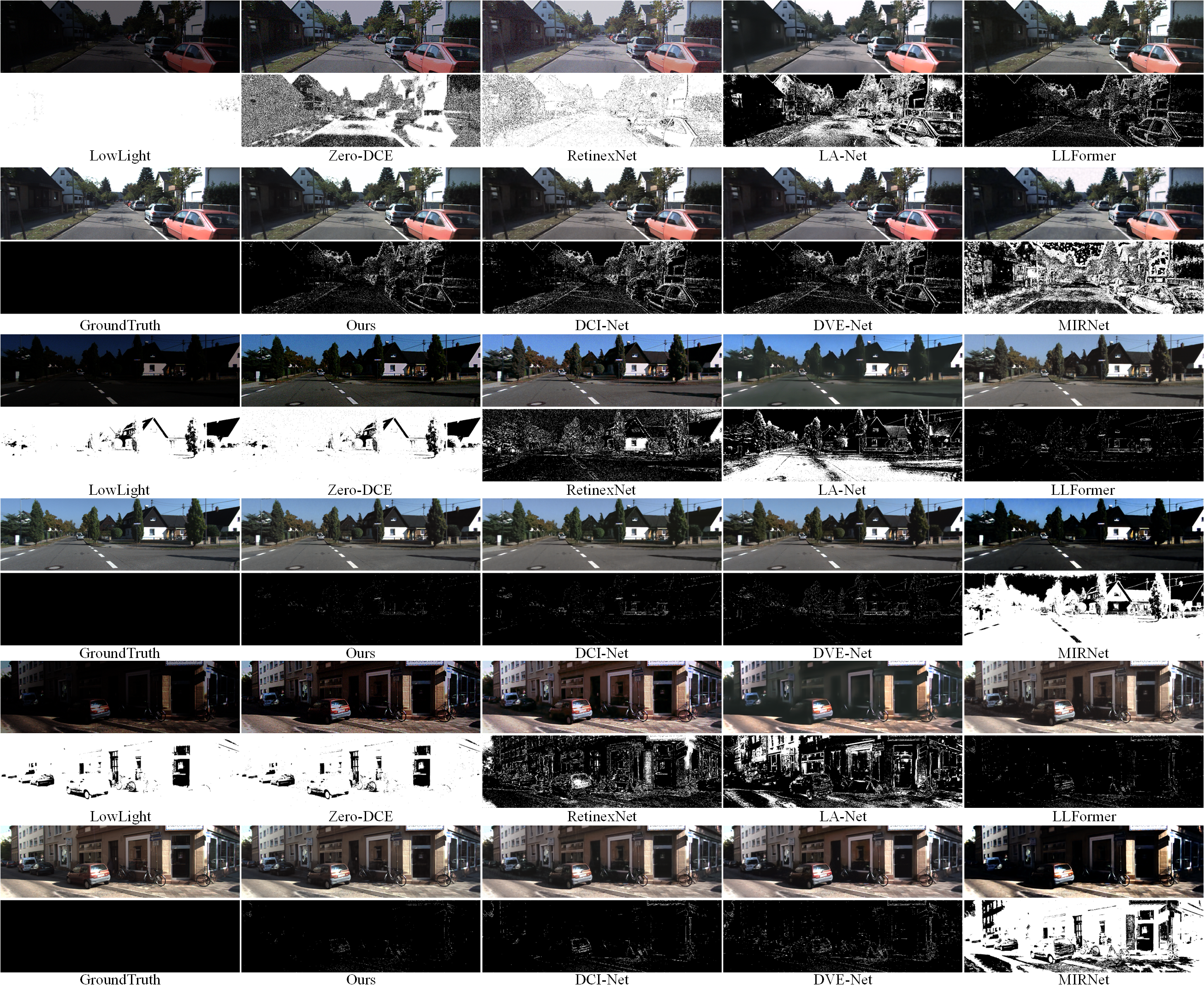}
	\caption{The MSE map between ground truth and the enhanced result on the Kitti2015 dataset, where the white area represents large error and the black area indicate small error. The results show that our method achieves the best enhancement results.}
	\label{pho:errormap}
\end{figure*}
\subsubsection{Implementation Details}
Our network is implemented with the Pytorch 2.0.1 framework and is trained with RTX 3090Ti GPU. During training, each patch is randomly cropped to 128 \(\times\) 128, with a batch-size of 20. On the validation set, we crop images to 400 × 400 and set the batch size to 1. Adam is used for optimization. The initial learning rate is set to 0.0002, and is reduced by half every 250 epochs. Our network has trained a total of 1000 epochs.

Additionally, in the low-frequency information enhanced module we use side window filter \cite{lowfre} to extract the low-frequency part of low-light images. Regarding the parameters, we set the radius to 5 and the iteration to 10. 
\subsubsection{Evaluation Metrics and Comparison Methods}
We employ two commonly used  full-reference image quality assessment metrics: PSNR and SSIM. Higher numerical values indicate better image quality. We compared our method with five single-image enhancement methods (Zero-DCE \cite{zero}, RetinexNet \cite{retinexnet}, LA-Net \cite{LA}, LLFormer \cite{llformer} and MIRNet \cite{mirnet}), and two stereo image enhancement methods (DVE-Net \cite{stereo-low2} and DCI-Net \cite{stereo-low3}). All methods used the same dataset as ours and were trained on the same GPU using the PyTorch framework. However, due to differences in network architectures, the hyper-parameters used may not be identical to ours.
\subsection{Quantitative Evaluations}
We conduct testing on three different datasets: Holopix50, Flickr2014, KITTI2015. The table \ref{tab:testdata} presents the PSNR and SSIM values for the images enhanced by our method and the compared methods. From the table, it can be observed that the unsupervised enhancement method Zero-DCE does not perform well.  The main reason is that there is no module designed for dealing with noise in the method, and it focuses on light and contrast adjustment. The single-view methods fall in the middle, as they can only extract information from a single view and cannot gather supplementary information from other views. This limitation may lead to suboptimal results in certain regions. In contrast, the stereo image enhancement methods achieve the better results. Among them, our method outperforms others in terms of both PSNR and SSIM, indicating superior image quality and structural improvements.
\begin{figure*}[t]\centering
	\includegraphics[width=7in]{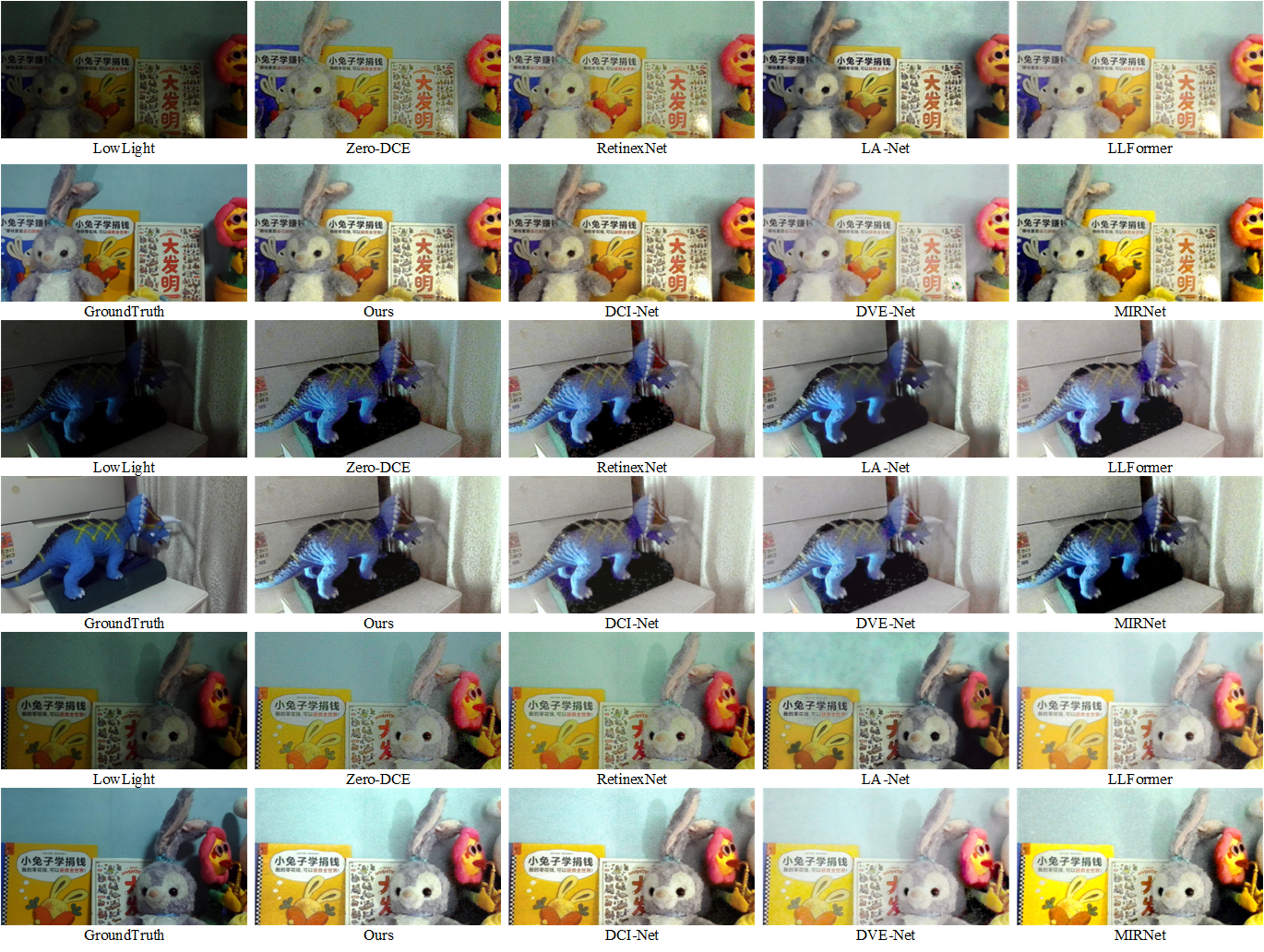}
	\caption{Visualization of the enhanced image of each method on Zed2 dataset. Compared to other methods, our method has better visualization results, but there is still a certain gap compared to ground truth. This is also the problem that we need to solve in our future work.}
	\label{pho:zed}
\end{figure*}

\subsection{Qualitative Evaluation}
Figure \ref{pho:holopix} and Figure \ref{pho:flickr} display enhanced images from the Holopix50 and Flickr2014 datasets. It is evident that our method performs well in color and structure restoration compared to other methods. Although some single view image enhancement methods perform equally well in color correction, there is still a gap in details recovery compared to stereo image enhancement methods. Compared with other stereo image enhancement methods, our results have fewer pseudo image shadow, better color, and are closer to ground truth.

Figure \ref{pho:errormap} shows the enhanced images and Mean Squared Error (MSE) maps between ground truth and the enhanced result on the Kitti2015 dataset, which are converted into binary image. The black regions represent smaller errors compared to the ground truth, while the white regions represent larger errors compared to the ground truth. It can be seen that our method achieves the lowest MSE error compared to the ground truth.

Figure \ref{pho:zed} shows the enhanced results on the Zed2 dataset. We found that all algorithms perform well in brightness adjustment, but not in noise suppression and image details recovery. There are two main reasons for this: 

(1) Images in the Zed2 dataset are captured in real scenes, and noise distribution in these images is very complex. While The low-light images in other datasets are generated with a  certain method. Thus, their noise distribution can be more easily learned and described by a latent space.

(2) In addition to low-brightness, low-contrast and noise, blur is a key degradation type in real low-light images. But we don't consider blur in synthetic dataset generation and training stage.

\subsection {Ablation Studies}
We present the results of the ablation studies in Table \ref{tab:ablation}.

(1) The impact of the IEM on performance: Table \ref{tab:ablation} shows that directly performing enhancement in the original low-light image space, enhancement performance will decrease. This indicate IEM can enhance the network’s denoising capability.

(2) The impact of diﬀerent feature interaction components on performance: W/o CVMI indicates removing the CVMI module, and W/o CSFI indicates removing the CSFI module. Table \ref{tab:ablation} shows the results on the Holopix50 dataset. If CVMI is missing, the network cannot capture features from another view. If CSFI is missing, the network has no interaction between different-scale features. We can see that removing any module will result in a decrease in performance, whether in terms of PSNR or SSIM.

(3) The impact of each loss function on performance: The results show that removing any one of the losses resulted in a decrease in network performance. By incorporating both losses, the reconstructed image quality is improved.

(4) The impact of each stages of CSM: CSM includes both spatial information mining and channel information mining stages, and it can be seen that removing either stage will result in a decrease in performance.
\begin{table}[]
	\small
	\newsavebox\CBox
	\def\textBF#1{\sbox\CBox{#1}\resizebox{\wd\CBox}{\ht\CBox}{\textbf{#1}}}
	\centering
	\caption{The table displays the the results of the ablation experiments. It can be found that any structure missing will lead to a decrease in performance and cannot achieve optimal results. In the table, W/o represent without the module. CSM-1 represents the first stage of the CSM, namely spatial information mining stage, and CSM-2 represents the second stage channel informationn mining module.}
	\begin{tablenotes}
		\item
	\end{tablenotes}
	\begin{tabular}{ccccc}
		\midrule[1pt] 
		\multirow{2}{*}{Module} & \multicolumn{2}{c}{Left}                            & \multicolumn{2}{c}{Right}                           \\ \cline{2-5} 
		& \multicolumn{1}{c}{PSNR} & \multicolumn{1}{c}{SSIM} & \multicolumn{1}{c}{PSNR} & \multicolumn{1}{c}{SSIM} \\ \hline
		W/o IEM            & 24.310             & 0.841             & 24.095             & 0.838             \\ \hline
		W/o CVMI                & 25.533             & 0.843             & 25.432             & 0.841             \\
		W/o CSFI                 & 24.178             & 0.828             & 24.026             & 0.826             \\
		W/o CVMI\&CSFI           & 22.342             & 0.821             & 22.075            & 0.817             \\ \hline
		W/o \(\mathcal{L}_{spa} \)                 & 25.760             & 0.853             & 25.597             & 0.850             \\
		W/o \(\mathcal{L}_{fre} \)                 & 25.447             & 0.855             & 25.270             & 0.853             \\ \hline
		W/o CSM-1                 & 25.753             & 0.859             & 25.524             & 0.811             \\
		W/o CSM-2                 & 25.367             & 0.854             & 25.197             & 0.846             \\ \hline
		Ours                    & \textbf{26.586}            & \textbf{0.868}             & \textbf{26.342}             & \textbf{0.860} \\
		\midrule[1pt]
	\end{tabular}
	\label{tab:ablation}
\end{table}
\section{Conclusion and Future Work}
We found that the existing low-light stereo image enhancement did not consider the impact of noise on cross-view interaction, and proposed a new low-light stereo image enhancement method. Specifically, our approach is to perform enhancement task in a new image space, which is obtained by fusing the original low-light image and its low-frequency part. This can reduce the impact of noise on feature encoding and interaction. Secondly, an encoder-decoder structure is used for image enhancement, and cross-view and cross-scale interactions are performed at multiple scales. In addition, we also propose a cross channel and spatial context information mining module to encode long-range spatial dependencies and enhance inter channel feature interaction. Sufficient experiments have shown that our method achieves the SOTA performance compared to existing methods. 

In our future work, we plan to address the challenges highlighted by the poor performance of existing methods on the Zed2 dataset. 
\section{Acknowledgments}
This research was funded by National Key Technology R\&D Program of China, grant number 2017YFB1402103-3, and Key Laboratory Foundation of Shaanxi Education Department, grant number 20JS086, National Natural Science Foundation of China (No.52275511).
\nocite{*}
\bibliographystyle{IEEEtran}
\bibliography{Low-light_Stereo_Image_Enhancement_and_De-noising_in_the_Low-frequency_Information_Enhanced_Image_Space.bib}
\end{document}